\documentclass[acmsmall, screen, authorversion, nonacm]{acmart}

\usepackage[ruled]{algorithm2e}
\usepackage{stmaryrd}
\usepackage{mathtools}
\usepackage{subcaption}
\DeclarePairedDelimiter{\Brackets}{\llbracket}{\rrbracket}
\newcommand{\R}{\mathbb{R}}
\newcommand{\unfold}{\operatorname{unfold}}
\newcommand{\diag}{\operatorname{diag}}

\usepackage{tikz}
\usetikzlibrary{shapes.geometric, arrows, positioning}

\tikzstyle{redrect} = [rectangle, rounded corners, 
minimum width=3cm, 
minimum height=1cm,
text centered, 
draw=black, 
fill=red!30]

\tikzstyle{bluecirc} = [circle, inner sep=0pt,minimum size=5pt, text centered, 
draw=black, fill=blue!30,
align=center]

\tikzstyle{circ} = [circle, inner sep=0pt,minimum size=5pt, text centered, 
draw=black]

\tikzstyle{orangerect} = [rectangle,
rounded corners,
minimum width=3cm, 
minimum height=1cm, 
text centered, 
text width=3cm, 
draw=black, 
fill=orange!30]

\tikzstyle{greenrect} = [rectangle,
rounded corners,
minimum width=3cm, 
minimum height=1cm, 
text centered, 
text width=3cm, 
draw=black, 
fill=green!30]

\tikzstyle{rect} = [rectangle,
rounded corners,
minimum width=1cm, 
minimum height=1cm, 
text centered,
align=center,
draw=black]

\tikzstyle{greendiam} = [diamond, 
minimum width=3cm, 
minimum height=1cm, 
text centered, 
draw=black, 
fill=green!30]
\tikzstyle{arrow} = [thick,->,>=stealth]

\AtBeginDocument{%
  \providecommand\BibTeX{{%
    \normalfont B\kern-0.5em{\scshape i\kern-0.25em b}\kern-0.8em\TeX}}}

\setcopyright{acmcopyright}
\copyrightyear{2023}
\acmYear{2023}
\acmDOI{10.1145/1122445.1122456}

\acmJournal{JACM}
\acmVolume{37}
\acmNumber{4}
\acmArticle{111}
\acmMonth{8}

\begin{document}

\title[Generating HMMs from Process Models Through Nonnegative Tensor Factorization]{Generating Hidden Markov Models from Process Models Through Nonnegative Tensor Factorization}

\author{Erik W. Skau}
\affiliation{%
  \institution{Information Sciences, Los Alamos National Laboratory}
  \country{USA}
}
\email{ewskau@lanl.gov}

\author{Andrew Hollis}
\affiliation{%
  \institution{Department of Statistics, North Carolina State University}
  \country{USA}
}
\email{anhollis@ncsu.edu}

\author{Stephan Eidenbenz}
\affiliation{%
  \institution{Information Sciences, Los Alamos National Laboratory}
  \country{USA}
}
\email{eidenben@lanl.gov}

\author{Kim \O. Rasmussen}
\affiliation{%
  \institution{Theoretical Division, Los Alamos National Laboratory}
  \country{USA}
}
\email{kor@lanl.gov}

\author{Boian S. Alexandrov}
\affiliation{%
  \institution{Theoretical Division, Los Alamos National Laboratory}
  \country{USA}
}
\email{boian@lanl.gov}

\renewcommand{\shortauthors}{Skau et al.}

\begin{abstract}
Monitoring of industrial processes is a critical capability in industry and in government to ensure reliability of production cycles, quick emergency response, and national security. Process monitoring allows users to gauge the progress of an organization in an industrial process or predict the degradation or aging of machine parts in processes taking place at a remote location.
Similar to many data science applications, we usually only have access to limited raw data, such as satellite imagery, short video clips, event logs, and signatures captured by a small set of sensors. To combat data scarcity, we leverage the knowledge of Subject Matter Experts (SMEs) who are familiar with the actions of interest. SMEs provide expert knowledge of the essential activities required for task completion and the resources necessary to carry out each of these activities. Various process mining techniques have been developed for this type of analysis; typically such approaches combine theoretical process models built based on domain expert insights with ad-hoc integration of available pieces of raw data. 
Here, we introduce a novel mathematically sound method that integrates theoretical process models (as proposed by SMEs) with interrelated minimal Hidden Markov Models (HMM), built via nonnegative tensor factorization. Our method consolidates: (a) theoretical process models, (b) HMMs, (c) coupled nonnegative matrix-tensor factorizations, and (d) custom model selection. To demonstrate our methodology and its abilities, we apply it on simple synthetic and real world process models.

\end{abstract}



\maketitle

\section{Introduction}
Process modeling, which is also called process mining, has been developed to analyze complex business enterprises that involve many people, activities, and resources to guide information systems engineering. Process models typically obtain their structure from workflow logs that describe past events relating to the enterprise process and specifications of how and which resources have been used~\cite{agrawal1998mining,van2007business}. 

When we monitor a specific process in real time, its activities and their temporal sequence are often not directly observable, and in this sense they remain hidden or latent. For instance, if we are monitoring an industrial process taking place at a remote/inaccessible location (e.g., building a industrial complex, such as an oil/liquefied gas terminal), we often have access to only a set of observables or indicators that underlie the activity, not the activity itself. Additionally, observable data is often scarce, as with remote sensing and event logs. Domain expert specification of the sequence of activities and their mean durations is useful to augment scarce data which can be used in statistical analysis. 

Process mining requires a statistical framework capable of accommodating both domain expert specifications and scarce observational data.
Given some series of observations from the process, this statistical framework should allow us to predict what process activity was underway at the time of the observations and how long it will be before the process is complete. Additionally, the statistical framework should be able to accurately evaluate the likelihood of competing domain expert process models based on limited observations. To answer these questions the statistical framework should describe the dynamics of the underlying activities (which are not directly observable) and how these hidden activities are related to the available observational data.

One such statistical framework is the Hidden Markov Model (HMM) introduced by Baum et al. \cite{baum1970maximization}. HMMs are a broadly used method for modeling data sequences and have been successfully applied in various fields, such as: signal processing, machine
learning, speech
recognition, handwritten character recognition, and in many other fields, see e.g., \cite{rabiner1986introduction,rabiner1989tutorial,song2001timing,apostolico2000optimal,perronnin2005probabilistic,cohen1998hidden,vaseghi2008advanced, rossi2015hmm, aceto2021characterization}. One of the well-known problems with HMMs is model selection, that is, how to estimate the optimal number of its hidden states. The HMM topology
and number of hidden states have to be known prior to its utilization, and various
heuristic rules have been proposed to estimate the HMM latent structure, see e.g., \cite{brand1998entropic}. Thus, the challenge inherent in using HMMs is that we must identify the minimal number of hidden states induced by the process model. We can then estimate the HMM parameters that describe the latent dynamics and the observational probabilities associated with each of the hidden states.

\begin{figure}
    \centering
    \begin{tikzpicture}[scale=0.75,node distance=3.75cm, every node/.append style={transform shape}]
    \node (process model) [orangerect] {Theoretical Process Model};
    \node (HMM) [redrect, below of=process model, node distance = 2.2cm] {HMM (not minimal)};
    \node (simulation) [orangerect, right of=process model] {Discrete Event Simulation};
    \node (tensors) [orangerect, right of=simulation] {Joint Probability Matrix and Tensor Construction};
    \node (decomposition) [orangerect, right of=tensors] {Coupled Nonnegative Matrix-Tensor Factorization with Model Selection};
    \node (minimal HMM) [greenrect, below of=decomposition, node distance=2.2cm] {Minimal HMM};
    \draw [arrow] (process model) -- (HMM);
    \draw [arrow] (process model) -- (simulation);
    \draw [arrow] (simulation) -- (tensors);
    \draw [arrow] (tensors) -- (decomposition);
    \draw [arrow] (decomposition) -- (minimal HMM);
    \end{tikzpicture}
    \caption{Flow chart illustrating the pipeline to construct a minimal HMM from a theoretical process model.}
    \label{fig:outline}
\end{figure}
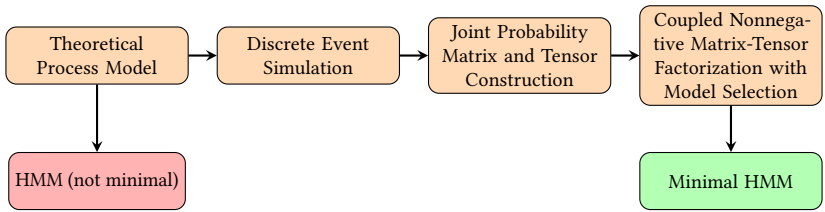

In this paper, we present a new method for building a minimal HMM based on a theoretical process model proposed by domain experts. Our method integrates: (a) theoretical process models, (b) HMMs, (c) coupled nonnegative matrix-tensor factorizations, and (d) custom model selection. The relative relations between these components can be seen in Figure~\ref{fig:outline}.

In summary, our contributions are:
\begin{itemize}
\item We demonstrate how to directly use the structure of a theoretical process model to build an HMM, where combinations of the process model activities are the HMM hidden states. One problem with this approach is the number of hidden states may not be minimal. Such misspecification can lead to a poor HMM parameterization, erroneous HMM predictions, and poor identifiability.
\item We show how to build an HMM model via coupled nonnegative matrix-tensor factorization, based on pairwise co-occurrence events \cite{huang2018learning, cybenko2011learning} generated from discrete event simulations \cite{schruben1993modeling} of a theoretical process model. 
\item Using the fact that tensor factorizations (and more specifically tensor ranks \cite{kruskal1977three}) can be used to derive the optimal (i.e., the minimal) number of HMM states \cite{allman2009identifiability, tune2013hidden}, we
utilize here our new method for Nonnegative Matrix Factorization (NMF) and Nonnegative Tensor Factorization (NTF) model selection \cite{alexandrov2020source,smarttensors2021}, and use it to determine the number of the hidden states based on the stability of the tensor decomposition. 
\item Finally, we demonstrate the results of our new method on synthetic and real world examples.
\end{itemize}

\section{RELATED WORKS}
In general, not much previous work exists that explores the connection between process modeling and HMMs. An HMM was proposed to sequence clustering in process mining \cite{da2009applying}. 
In \cite{carrera2014constructing}, the authors constructed an HMM for resource allocation, and used an HMM to identify
the observations based on the resources utilized by each activity. HMMs have also been used for the representation of process mining of parallel business processes \cite{sarno2016hidden}. Similarly, HMMs were used for extraction of activity information recorded in the bank event log, for calculating the fraud probabilities \cite{rahmawati2017fraud} and for detection of anomalies in business processes \cite{yang2020analysing}. The authors of \cite{jaramillo2019automatic} applied HMMs for business process management mining for failure detection from event
logs. In the area of healthcare and medicine, HMMs were used for complex healthcare process analysis and detection of hidden activities \cite{alharbi2019unsupervised}. Recently it was demonstrated how to build a semi-Markov model based on event sequences, \cite{kalenkova2022performance}, and to use it for automatic activity discovery and analysis. Finally, although there are several works relating minimal HMMs and tensor factorization (see for example \cite{ohta2021realization} and references therein), to the best of our knowledge, our work is the first one that explores the connection between theoretical process modeling and tensor factorization. There are numerous works highlighting the importance of accurate tensor factorizations and the difficulty of nonnegative tensor factorizations~\cite{kolda2009tensor,luo2019temporal, wu2020advancing, wu2021pid}. In this work we simultaneously decompose matrices and tensors with a joint least squared objective function. This type of minimization objective is often called coupled matrix-tensor factorization~\cite{acar2011all, schenker2021optimization}, though every instance requires slightly different formulations as the factorization structures vary between applications.

One necessity of HMMs is the determination of the latent dimension. When using matrix or tensor factorization to derive HMMs, this dimension is called the rank and is typically denoted with "k". The rank is typically elusive in practical applications, demanding an estimation derived exclusively from the data. Despite the absence of a polynomial-time algorithm to ascertain a tensor's rank - a task proven to be NP-hard~\cite{johan1990tensor} - various heuristic methodologies have been developed for nonnegative rank estimation.
One prominent technique is the core consistency diagnostic, known as CORCONDIA~\cite{bro2008cross}. This method evaluates deviations from a super-diagonal core-tensor, estimating the number of components, k, where this deviation reaches a minimum. Despite its intuitive appeal and widespread application, CORCONDIA lacks a solid theoretical foundation. Particularly for nonnegative tensor factorization, it encounters challenges when contrasting a nonnegative Tucker Decomposition with the corresponding nonnegative Canonical Polyadic Decomposition (CPD). For the complications of these two nonnegative decompositions see~\cite{alexandrov2022nonnegative}.
An alternative strategy is the Bayesian approach, incorporating Automatic Relevance Determination (ARD). Initially devised for neural networks~\cite{mackay1994automatic} and subsequently adapted for Bayesian PCA~\cite{bishop1998bayesian}, and then to NMF~\cite{tan2012automatic, morup2009tuning}, and tensors~\cite{morup2009automatic}, ARD is an alternative approach to identify the number of latent dimensions.
Other strategies employ diverse statistical criteria, including minimum description length~\cite{liu2016detection}, Akaike Information Criterion (AIC)~\cite{sakamoto1986akaike}, and Bayesian Information Criterion (BIC)~\cite{schwarz1978estimating} with partial success. Determining the optimal nonnegative model - that is, the nonnegative rank - remains a formidable challenge, with varying approaches often leading to disparate solutions~\cite{hansen1999generalizable}.
The methodology employed in our study hinges on the well-posed nature of the low-rank approximation of the nonnegative CPD, where the solutions are almost always unique~\cite{qi2016uniqueness}. We determine the number of latent components, k, by analyzing the stability of the solutions from factorizations across a set of plausible k values ~\cite{brunet2004metagenes}. Our method demonstrated superior efficacy when tested on numerous synthetic datasets with pre-established latent features~\cite{nebgen2021neural}. Additionally, our approach leverages the theorem that a unique NMF solution, for a matrix: X = WH, when X is subjected to minor noise or perturbations, results in minimal disruptions to the factors W and H~\cite{laurberg2008theorems}. 
 
Our method has seen successful application in various domains, including the decomposition of the largest human cancer genome data~\cite{alexandrov2013signatures}, synthetic cancer data~\cite{islam2022uncovering}, microphase separation of block copolymers, via Nonnegative CPD~\cite{alexandrov2019nonnegative}, analysis of international trade flows~\cite{truong2020determination} and exabyte synthetic data~\cite{bhattarai2023distributed}, via RESCAL, Nonnegative Tucker decompositions for images~\cite{pulido2021selection} as well as for large set of trajectories of the reaction diffusion equation~\cite{vesselinov2019unsupervised}, among others. Our methodology has also been utilized for model selection in Boolean~\cite{desantis2022factorization}, Quaternonian decompositions~\cite{sanchez2021automatic}, and not-multilinear decompositions\cite{stanev2018identification, vangara2020identification}. 

\section{Process Modeling, Hidden Markov Models, and nonnegative Tensor Factorization with Model Selection}

\subsection{Theoretical Process Modeling}
\label{sec:process_model}
Constructing a theoretical process model requires specifying the set of activities making up the process, the order in which these activities take place, the expected duration of each activity, and the resources necessary to execute each activity. In this endeavor, we rely on subject matter expertise to craft the theoretical process model that incorporates insights and observations provided by experts knowledgeable about the process. We utilize the discrete event simulation tool, Simian simulation engine \cite{santhi2015simian}, to actively simulate a process model. 

Simian relies on three main components to define a process model: the precedence constraints, the activity duration distributions, and the required resources for each activity. Simian takes a Directed Acyclic Graph (DAG) $G = (V,E)$ to define the precedence constraints, where the vertex set $V = \{a_1, \ldots, a_n\}$ represent activities of the process, and the edge set $E \subseteq V^2$ represent precedent constraints. For example, if edge $(a_i, a_j) \in E$, activity $a_i$ must be have been completed before process $a_j$ can be started. Each vertex in the DAG has a distribution of feasible activity durations. Each simulation picks values from the distribution for the duration of each activity $a_i$. We constrain our models to have exponentially distributed durations, so the duration of activity $a_i \sim \frac{1}{\beta_i} e^{-x/\beta_i}$. Simian also requires the specification of a set of resources $R_i \subseteq R$  required for the execution of each activity, where $R = \{r_1, \ldots, r_m\}$ is a global set of resources. In our study, we assign an infinite quantity of each resource to $R$, ensuring that limited resources will never impede an activity from starting. Additionally, we assign a probability that each resource will be observed for each combination of activities simulating the stochastic nature of realistic scenarios. We utilize the resources necessary for each activity as observable quantities.

A Simian simulation ensures compliance with the activity precedence's specified by the DAG, assigns durations to each individual activity, and verifies the availability of required resources for each activity to begin. In each simulation, Simian provides data regarding the start and end times of each activity. This information can be organized to create timelines of both the unobservable process model activities and the observable quantities of resource usage. For complex process models, every simulation run by Simian will have varying durations for each activity, resulting in diverse combinations of ongoing activities throughout the process. That is to say, the details of which activities run in parallel, and for how long, will change from simulation to simulation due to the sampled duration choices. Thus, we resort to statistical analysis on large ensembles of simulations to calculate probabilities of transitioning between each resource usage state. For our statistical analysis we join a collection of Simian simulationss so that the end of each simulation leads into the beginning of the next simulation, effectively making the process model repeat \emph{ad infinitum}.

%


\subsection{Hidden Markov Models}

\begin{figure}
    \centering
    \begin{tikzpicture}[node distance=.75cm and .75cm,>=stealth',thick]
  \newcommand\Ellipse[3][]{\node[circle,minimum size=1.1cm,draw,inner sep=2pt,#1](#2){#3};}
  \newcommand\hidden[2][]{\Ellipse[#1]{hidden #2}{$\mathit{X}_{#2}$}}
  \newcommand\observed[2][]{\Ellipse[#1]{observed #2}{$\mathit{Y}_{#2}$}}
  \hidden{t-1};
  \node[left=of hidden t-1,minimum width=1cm] {$\pi$};
  \hidden[right=of hidden t-1]{t};
  \hidden[right=of hidden t]{t+1};
  \observed[below=of hidden t-1]{t-1};
  \observed[right=of observed t-1]{t};
  \observed[right=of observed t]{t+1}
  
  \draw[->] (hidden t-1) edge node[right,midway]{$E$} (observed t-1) edge node[above,midway]{$T$} (hidden t)
            (hidden t) edge node[right,midway]{$E$} (observed t) edge node[above,midway]{$T$} (hidden t+1)
            (hidden t+1) edge node[right,midway]{$E$} (observed t+1);
  \draw[->, dashed] (hidden t+1) edge +(1.5,0);
  \draw[<-, dashed] (hidden t-1) edge +(-1.5,0);
    \end{tikzpicture}
    \caption{Depiction of an HMM with hidden random variable $X$ and observed random variable $Y$. $pi$ depicts the initial state probability vector. The transition matrix $T$ describes the progression of the hidden random variable at time $t$ to time $t+1$. The emission matrix $E$ describes the probability of each observation based upon the hidden state at each time.}
    \label{fig:HMM}
\end{figure}
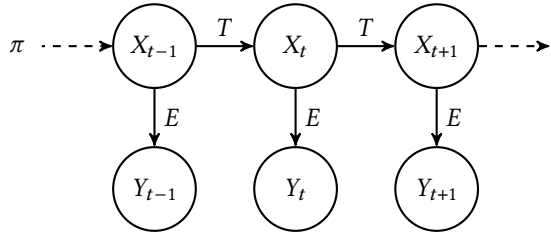
HMMs offer a popular statistical framework for modeling processes that are not directly observable but for which we can observe related data at discrete time intervals. The observable data is usually various objects or events we can detect. For instance, if we are in a windowless building, we may not be able to see what the weather is like outside, but we can observe someone carrying an umbrella or wearing sunglasses at different times. 

Mathematically, an HMM has a set of $N$ hidden states, each of which emits observations with given probabilities. The HMM is defined by three parameters: the transition matrix, the emission matrix and the initial state probability vector. The $N \times N$ transition matrix $T$ and length $N$ initial state vector $\pi$ fully specify the underlying Markov chain. The transition probability matrix, $T$, gives the probabilities of transitioning from one hidden state to another. The transition matrix incorporates information about the duration of states and determines the order of states by specifying how the process transitions between them. We consider the probabilities of transitioning at discrete homogeneous time intervals. The initial probability vector, $\pi$, specifies the probabilities that the process will be in one of the $N$ hidden states when the observations begin. The last key parameter of HMM is the emission probability matrix, $E$, that connects the observations with the hidden states. Assuming there are $M$ possible observations, $E$ is an $M \times N$ matrix that specifies the probability one of the $M$ observation configurations is emitted during a given activity (which is one of the $N$ hidden states). Figure ~\ref{fig:HMM} depicts the transition and emission relations in an HMM.


\subsection{Creating Reference Hidden Markov Models from a Theoretical Process Model}
\label{sec:reference_hmm}

While in general, a theoretical process model is not a perfect fit with HMMs, a relation between groups of simultaneous activities in a process model and hidden states in an HMM can immediately be drawn. For a given number of activities in a theoretical process model we can upper bound the number of states in an associated HMM. Given a theoretical process model with $n$ activities, the corresponding HMM will have at most $2^n$ states so $N \leq 2^n$. This bound is obtained when a process model has $n$ activities in parallel. In this case, every subset of activities can be achieved through different combinations of activity durations, generating the power set of the $n$ process model activities. Similarly, with $m$ resources, there will be at most $2^m$ combinations of various resource usage so $M \leq 2^m$.

The additional restrictions detailed in Section~\ref{sec:process_model} strengthen the connection between process models and HMMs. Namely, having exponentially distributed activity durations, in conjunction with having infinite resources, provides the memory-less Markovian property. These restrictions make the probability of transitioning from one hidden state to another independent of all previous history. Joining multiple Simian runs together corresponds to a repeating process model, and sampling them throughout at discrete intervals, provides sampling corresponding to an ergodic HMM. With this joining, every state can transition to every other state given enough time. With these restrictions, a correspondence between a theoretical process model and an ergodic HMM is established. 

It is simple to obtain empirical estimates of HMM parameters by simulating the process model and accumulating statistics from discrete time samplings of simulations in a sequential manner. Suppose we wish to model an industrial process with $n$ activities $\{a_1,...,a_n\}$ and $m$ resources $R = \{r_1,...,r_m\}$. The HMM hidden states, $\{s_1, \hdots, s_N\}$ with $N \leq 2^n$, consist of all combinations of activities present in all simulations. The HMM observed states, $\{h_1, \hdots h_M\}$ with $M \leq 2^m$, enumerate all combinations of resources utilized in all simulations. We take $\pi_i$ to be the proportion of times the simulated data stream is in state $s_i$. $T_{ij}$ is estimated as the proportion of times the simulations transitioned from $s_i$ to $s_j$. $E_{ij}$ is taken to be the proportion of times observation type $h_j$ is seen while the process is in state $s_i$ across all simulated runs. 

While this can always be done empirically, regardless of restrictions forcing a process model to adhere to our HMM assumptions, in nice cases an HMM can be constructed from theoretic principals. Using the mean activity completion times $\{\beta_1,...,\beta_n\}$ the HMM parameters $\pi$, $T$, and, $E$, can be derived using basic properties of continuous time Markov chains. Let $\textbf{Q}$ be am $N \times N$ matrix called the transition rate matrix. For $i \neq j$, the $ij$-th entry of $\textbf{Q}$ is the inverse of the mean duration of the activity that causes state $i$ to transition to state $j$. For $i=j$, the $ii$-th entry is the negative of the sum of the other entries in row $i$. It can be shown that $T$ can be written as a matrix exponential \citep{ross1996stochastic}:

\begin{equation}
\label{eqn:ref_hmm}
T=e^{\textbf{Q}(\Delta t)}=\sum_{n=0}^\infty \frac{1}{n!}(\textbf{Q}(\Delta t))^n\;,
\end{equation}
where $\Delta t$ is the discrete time interval of interest.
In general, we cannot analytically compute matrix exponentials, but standard numerical approximations exist \citep{higham2005scaling, moler2003nineteen, sastre2015new}. 

An HMM built in this way could precisely quantify the observable dynamics of the process, but this HMM representation is not unique. There are equivalency classes of HMMs, and two HMMs whose output processes are statistically indistinguishable belong to the same equivalency class. Due to the families of equivalent HMMs the determination of a 'true' HMM is an ill posed problem. Often, minimal HMMs are sought within the model equivalency class~\citep{huang2018learning}. Preferring HMMs with a minimal number of hidden states results in stable parameter estimations from limited or scarce data, and provides clearer and explainable results. Matrix and tensor factorizations are one of the successful methods in finding minimal HMMs~\citep{huang2015minimal}.

\subsection{Nonnegative Tensor Factorization and Model Selection}
\label{sec:tensor_factorization}
We employ a coupled nonnegative matrix-tensor factorization method to estimate HMM parameters. The coupled factorization simultaneously decomposes a joint probability tensor, $P(Y_{t-1}, Y_t, Y_{t+1})$, and joint probability matrix, $P(Y_t, Y_{t+1})$. It is simple to empirically construct these arrays by counting the number of occurrences of each sequence of events from many Simian runs. The $(i,j,k)$ entry of the the tensor $P(Y_{t-1}, Y_t, Y_{t+1})_{i,j,k}$ is the proportion of number of occurrences of the $(i,j,k)$ sub-sequence in the observed resources used across all Simian runs. Similarly, $P(Y_t, Y_{t+1})_{i,j}$ is the normalized number of occurrences of the (i,j) sub-sequence. The joint probability tensor can be approximated with a nonnegative CPD \cite{huang2018learning, chi2012tensors},
\begin{multline}
    P(Y_{t-1} = h_i, Y_t = h_j, Y_{t+1} = h_k) \approx\\ \sum_{n=1}^{N} P(X_t = s_n) P(Y_{t-1} = h_i | X_t = s_n) P(Y_{t} = h_j | X_t = s_n) P(Y_{t+1} = h_k | X_t = s_n)\;,
\end{multline}
where $P(Y_{t} | X_t = s_n)$ is the emission matrix for an $N$ hidden state Markov model. The joint probability matrix can similarly be decomposed 
\begin{equation}
    P(Y_t=h_i, Y_{t+1}=h_j) \approx \sum_{i=1}^{N} \sum_{j=1}^{N} P(Y_{t} = h_i | X_t = s_i) P(X_t = s_i, X_{t+1} = s_j) P(Y_{t+1} = h_j | X_t = s_j)\;,
\end{equation}
where $P(X_t = s_i, X_{t+1} = s_j)$ is a joint probability matrix relating the transitions of the $N$ hidden states. To make notation more manageable we apply a change of variables,

\begin{align*}
    \mathcal{T}_{i,j,k} &= P(Y_{t-1} = h_i, Y_t = h_j, Y_{t+1} = h_k)\\
    M_{i,j} &= P(Y_t = h_i, Y_{t+1} = h_j)\\
    A_{i,j} &= P(X_t = s_j) P(Y_{t-1} = h_i | X_t = s_j)\\
    B_{i,j} &= P(Y_{t} = h_i | X_t = s_j)\\
    C_{i,j} &= P(Y_{t+1} = h_i | X_t = s_j)\\
    D_{i,j} &= P(X_t = s_i, X_{t+1} = s_j)\;,
\end{align*}
to concisely express the tensor and matrix approximations,
$$\mathcal{T} \approx \Brackets{A,B,C},\quad M \approx B D B^\top\;,$$
where $\Brackets{A,B,C}_{i,j,k} = \sum_{n=1}^{N} A_{i,n} \cdot B_{j,n} \cdot C_{k,n}$.
To identify viable HMM models consistent with the joint probability tensor and matrix, we solve a coupled nonnegative matrix-tensor optimization problem,
\begin{equation}
\begin{aligned}
    \underset{A,B,C,D}{\operatorname{minimize}} \quad & \frac{1}{2}\Vert \mathcal{T} - \Brackets{A,B,C}\Vert_F^2 + \frac{\rho}{2} \Vert M - B D B^\top \Vert_F^2\\
    \textrm{subject to} \quad & A_{i,j} \geq 0, \quad \sum_{i,j} A_{i,j} = 1 \\
    & B_{i,j} \geq 0, \quad \sum_{i} B_{i,j} = 1 \text{ for } 1 \leq j \leq N \\
    & C_{i,j} \geq 0, \quad \sum_{i} C_{i,j} = 1 \text{ for } 1 \leq j \leq N \\
    & D_{i,j} \geq 0, \quad \sum_{i,j} D_{i,j} = 1
\end{aligned}
\label{eq:optimization}
\end{equation}
where, $\rho$ is a regularization parameter controlling the relative importance of the two terms, the resulting matrix $B$ corresponds to the emission matrix, and the transition matrix can be derived from $D$ by normalizing the rows to sum to 1.

To solve the optimization we employ the multiplicative updates method. Multiplicative updates are one of many optimization techniques for nonnegative matrix and tensor factorization in the company of, BFGS-B, alternating direction method of multipliers~\cite{sun2014alternating}, and many others~\cite{kim2014algorithms}. While multiplicative updates do not always have be best convergence, multiplicative updates are simple to implement and adapt to our nonlinear properties of $B$ in $M \approx B D B^\top$. Multiplicative updates alternates updating the various matrices as block coordinates and leverages that nonnegative numbers are closed under multiplication to enforce the nonnegative constraint. To minimize our objective function $J(A,B,C,D) = \frac{1}{2}\Vert \mathcal{T} - \Brackets{A,B,C}\Vert_F^2 + \frac{\rho}{2} \Vert M - B D B^\top \Vert_F^2$, multiplicative updates iterates over $A,B,C,D$ and element-wise multiplies each one with the ratio of the negative and positive components of the partial derivative. To this end, we compute the partial derivatives of the objective function with respect to each individual component matrix:
\begin{equation}
    \begin{aligned}
        \cfrac{\partial{J}}{\partial{A}} &= A \left( (C^\top C) \odot (B^\top B) \right) - \unfold(\mathcal{T},1) (C \otimes B)\\
        \cfrac{\partial{J}}{\partial{B}} &= B \left( (A^\top A) \odot (C^\top C) \right) - \unfold(\mathcal{T},2) (C \otimes A) + \rho \left( B D^\top B^\top B D + B D B^\top B D^\top - M B D^\top - M^\top  B D \right)\\
        \cfrac{\partial{J}}{\partial{C}} &= C \left( (B^\top B) \odot (A^\top A) \right) - \unfold(\mathcal{T},3) (B \otimes A)\\
        \cfrac{\partial{J}}{\partial{D}} &= B^\top B D B^\top B - B^\top M B\;,
    \end{aligned}
\end{equation}
where $\otimes$ is the Khatri-Rao product \cite{khatri1968solutions}, and $\odot$ the Hadamard product \cite{million2007hadamard}.
With the nonnegative property of all the factors, all components with negative signs constitute the negative component of the partial derivative, and the remaining terms the positive component. The probability constraints of summing to 1 are applied as post-processing by applying appropriate scaling to each of the factors. The resulting alternating optimization algorithm is displayed in Algorithm~\ref{alg:joint_optimization}.
\begin{algorithm}
\caption{Joint Optimization Algorithm}\label{alg:joint_optimization}
\KwData{$\mathcal{T} \in \R^{K,K,K}_+, M \in \R^{K,K}_+, A \in \R^{K,N}_+, B \in \R^{K,N}_+, C \in \R^{K,N}_+, D \in \R^{N,N}_+$}
\KwResult{$\mathcal{T} \approx \Brackets{A,B,C}, M \approx B D B^\top$}
\While{not converged}{
    $A \gets A \cfrac{\unfold(\mathcal{T},1) (C \otimes B)}{A \left( (C^\top C) \odot (B^\top B) \right)}$\;
    $B \gets B \cfrac{\unfold(\mathcal{T},2) (C \otimes A) + \rho \left( M B D^\top + M^\top  B D \right)}{B \left( (A^\top A) \odot (C^\top C) \right) + \rho \left( B D^\top B^\top B D + B D B^\top B D^\top \right) }$\;
    $C \gets C \cfrac{\unfold(\mathcal{T},3) (B \otimes A)}{C \left( (B^\top B) \odot (A^\top A) \right)}$\;
    $D \gets D \cfrac{B^\top M B}{B^\top B D B^\top B}$\;
}
$ A \gets A \diag(\sum_{i} B_{i,j}) \diag(\sum_i C_{i,j}) $\;
$ A \gets \frac{A}{\sum_{i,j} A_{i,j}} $\;
$ D \gets \diag(\sum_{i} B_{i,j}) D \diag(\sum_{i} B_{i,j})$\;
$ D \gets \frac{D}{\sum_{i,j} D_{i,j}}$\;
$ B \gets B \diag(\sum_{i} B_{i,j})^{-1} $\;
$ C \gets C \diag(\sum_{i} C_{i,j})^{-1} $\;

\end{algorithm}

\begin{figure}
    \centering
\begin{tikzpicture}[scale=0.55,node distance=.5cm, every node/.append style={transform shape}]
    \node[rect] (input) at (0,0)              {Input:\\
    Tensor $\mathcal{T}$\\
    Matrix $M$};
    
    \node[rect, right=of input] (ensemble) {Resampled Ensemble:\\
    $\{\mathcal{T}_1, \mathcal{T}_2, \hdots, \mathcal{T}_r\}$\\
    $\{M_1, M_2, \hdots, M_r\}$};
    
    \node[rect, right=of ensemble] (factor) {Suppose $k$ latent dimensions and factor resampled arrays $\mathcal{T}_q$, $M_q$\\
    $A_q, B_q, C_q, D_1 = \underset{A,B,C,D}{\operatorname{minimize}} \quad \frac{1}{2}\Vert \mathcal{T} - \Brackets{A,B,C}\Vert_F^2 + \frac{\rho}{2} \Vert M - B D B^\top \Vert_F^2$\\
    for $1 \leq q \leq r$};
    
    \node[rect, right=of factor] (emission)
    {Collect Factors\\
    $\{B_1, \hdots, B_r\}$};
    
    \node[rect, right=of emission] (cluster)  {Cluster factors\\into $k$ clusters\\\\\begin{tikzpicture}[scale=1]%
        \node[circ, minimum size=.65cm](c1) at (-0.6495,0.375) {\#1};
        \node[circ, minimum size=.65cm](c2) at (0.155925, 0.7335) {\#2};
        \node[circ, minimum size=.65cm](c3) at (0.74625, 0.078) {\#3};
        \node[circ, minimum size=.65cm](dots) at (0.30524999999999997, -0.6855) {$\hdots$};
        \node[circ, minimum size=.65cm] (ck) at (-0.55725, -0.5025) {\#k};
   \end{tikzpicture} };
   
   \node[rect, right=of cluster] (stability) 
    {Check Cluster\\Stability};
    
    \draw[arrow] (input) -- (ensemble);
    \draw[arrow] (ensemble) to (factor);
    \draw[arrow] (factor) to (emission);
    \draw[arrow] (emission) to (cluster);
    \draw[arrow] (cluster) to (stability);
    \draw[arrow] (stability.south) -- +(0,-1.5) -| (factor.south) node [pos=.25, above] {$k = k + 1$};

\end{tikzpicture}
    \caption{Diagram of the steps taken to construct a minimal HMM.}
    \label{fig:model_selection}
\end{figure}
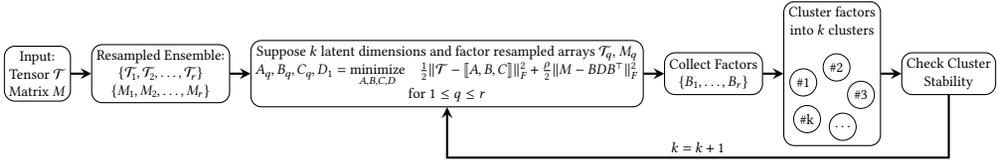

To determine the optimal/minimal number of hidden states, $N$, we utilize our method that relies on the stability of factors from slightly perturbed arrays~\cite{alexandrov2019nonnegative}. Specifically, we: a) create an ensemble of randomly perturbed joint probability tensors and matrices, b) decompose these ensembles, and c) cluster and measure the stability of the resulting set of factors. A diagram depicting this procedure is in Figure~\ref{fig:model_selection}. The ensemble of perturbed arrays is generated by element-wise perturbing each component of the arrays $X$ as $X_{\text{perturbed}} = X \odot \mathcal{U}(1-\epsilon, 1+\epsilon)$ where $\mathcal{U}(a,b)$ is the uniform distribution on the interval $[a,b]$. Each tensor and matrix pair of this random ensemble is decomposed by solving the optimization in Equation (\ref{eq:optimization}) for each candidate latent dimension, $k = 1,2,...,N-1$. For each candidate latent dimension the emission matrix factors are clustered to remove the permutation ambiguity of the decompositions. We apply the clustering to the emission matrix, $B = P(Y_{t} | X_t),$ as it directly appears in both terms of the objective function and will be identifiable from the tensor decomposition. The clustering procedure, previously reported in \cite{vangara2020semantic,nebgen2021neural}, is similar to applying k-means to the vectors of the emission matrices. The only difference is our custom clustering requires each cluster to contain exactly one vector from each emission matrix in the ensemble. Finally after the clustering, silhouettes statistics \cite{rousseeuw1987silhouettes} are used to score the quality of the clusters and the objective function is used to measure the quality of the decomposition. Silhouette scores measure the intra-cluster distance compared to the nearest-cluster distance and vary from $[-1,1]$ with a value of $1$ indicating a perfect clustering and $-1$ indicating very poor clustering. The objective function indicates how well the decompositions represent the data with values close to $0$ indicating a better representation. A suitable number of hidden variables is selected from the candidate dimensions as the candidate with a high average silhouette score and low objective function value. In practice, we test several candidate hidden dimensions $1 \leq k \leq N-1$ and select a suitable latent dimension based on cluster stability and fit criteria. 

\section{Numerical Experiments}

Numerically constructing a minimal HMM from a given theoretical process model means finding the transition and emission matrices of an HMM with a minimal number of hidden states. After we build a minimal HMM, we can use standard HMM algorithms \cite{rabiner1986introduction} to find from a sequence of observations: (i) What sequence of states likely produced this observation sequence; (ii) What state the process is likely in after we have made our observations; and (iii) The likelihood that the sequence of observations came from that HMM derived from the process model. Below we demonstrate examples of how to numerically build a minimmal HMM from theoretical process models via tensor factorization with estimations of the minimal number of states.

To evaluate our procedure to construct minimal HMMs, we used Simian to simulate 1,000,000 runs for each process model and sampled the active processes and observed resources at 20 hour intervals unless otherwise specified. Our objective function evaluations were carried out with the regularization parameter $\rho=1$, which resulted in approximately equal contributions from the matrix and tensor residuals to the objective function. Datasets with more disparate norms between the arrays, or disparate sizes require careful consideration of this metaparameter. From the sequence of observed resources we construct the joint probability tensor and matrix empirically following the procedure in Section~\ref{sec:tensor_factorization}. Perturbed ensembles of the tensor and matrix are jointly decomposed with Algorithm~\ref{alg:joint_optimization} to recover the optimal emission and unnormalized transition matrices for each candidate latent dimension. The clustering and latent dimension selection method, described in Section \ref{sec:tensor_factorization} is applied for ensembles of $r=1000$ perturbations, but the cluster stability is only evaluated on the 10\% of solutions with the lowest objective function to remove poor fits.

For evaluation, we utilized reference HMMs constructed from the process model as described in Section~\ref{sec:reference_hmm}. We simulated $T$ long runs of the hidden states, $X_{1:T}^{ref}$, and corresponding observations, $Y_{1:T}^{ref}$, from the reference HMM and compared the performance of each HMM at predicting the sequence of hidden states, $X_{1:T}^{pred}$, using the Viterbi \cite{viterbi1967error} algorithm on the sequence of observations. As the reference HMM and Tensor Factorization (TF) HMMs often have different numbers of hidden states, we evaluate each model with normalized mutual information,
$$
\operatorname{nmi}(X_{1:T}^{ref}, X_{1:T}^{pred}) = \frac{D_{KL}( P(X_{1:T}^{ref},X_{1:T}^{pred}) \vert P(X_{1:T}^{ref}) \otimes P(X_{1:T}^{pred})}{H(X_{1:T}^{ref}) + H(X_{1:T}^{pred})}
$$
where $D_{KL}$ the Kullback-Leibler divergence, and $H$ the entropy. We additionally report the distance,
$$\operatorname{dist}(\lambda, \hat{\lambda}) = \mathbb{E} \left[ \frac{1}{T}(\log P_{\lambda}(Y_{1:T}^{ref}) - \log P_{\hat{\lambda}}(Y_{1:T}^{ref})) \right]\;,$$
between the best HMM with $k$ hidden states, $\lambda$, relative to the best HMM with $k-1$ hidden states, $\hat{\lambda}$, from the reference observed sequences.

\subsection{Synthetic Example with Process Model Activities Leading to a Minimal HMM}

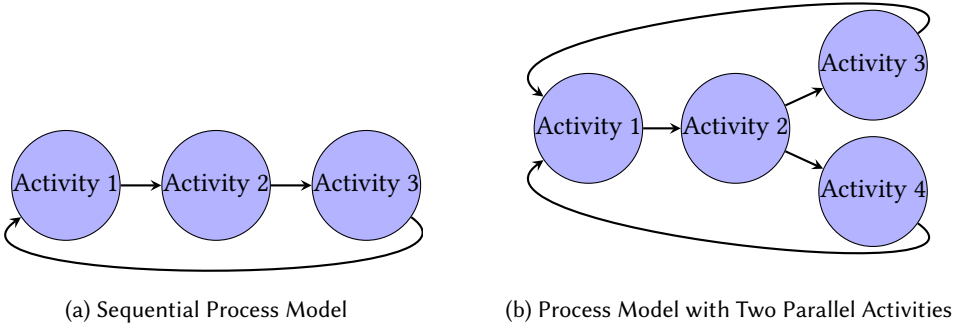
\begin{figure}
\centering
\begin{subfigure}{.5\textwidth}
  \centering
  \begin{tikzpicture}[node distance=2cm]
    \clip (-1,-1.25) rectangle (4.75,.75);
    \node (1) [bluecirc] {Activity 1};
    \node (2) [bluecirc, right of=1] {Activity 2};
    \node (3) [bluecirc, right of=2] {Activity 3};
    \draw [arrow] (1) -- (2);
    \draw [arrow] (2) -- (3);
    \draw [arrow] (3) to[out=-35, in=-145, looseness=.8] (1);
\end{tikzpicture}
\caption{Sequential Process Model}
\label{fig:serial_pm}
\end{subfigure}%
\begin{subfigure}{.5\textwidth}
  \centering
  \begin{tikzpicture}[node distance=0cm and .5cm]
    \clip (-1,-2) rectangle (4.75,2);
    \node (1) [bluecirc] {Activity 1};
    \node (2) [bluecirc, right=of 1] {Activity 2};
    \node (3) [bluecirc, above right=-.2cm and .8cm of 2] {Activity 3};
    \node (4) [bluecirc, below right=-.2cm and .8cm of 2] {Activity 4};
    \draw [arrow] (1) -- (2);
    \draw [arrow] (2) -- (3);
    \draw [arrow] (2) -- (4);
    \draw [arrow] (3) to[out=35, in=145, looseness=.8] (1);
    \draw [arrow] (4) to[out=-35, in=-145, looseness=.8] (1);
\end{tikzpicture}
\caption{Process Model with Two Parallel Activities}
\label{fig:concurrent_pm}
\end{subfigure}
\caption{Two example process models where the final activities connect to the first activity causing the process models to repeat.}
\label{fig:test}
\end{figure}

First we analyzed a small process model, depicted in Figure \ref{fig:serial_pm}, with no parallelism. We demonstrate that the reference HMM and the minimal HMM obtained by tensor decomposition produce the same results. \\

The reference HMM can be constructed with mean durations for each activity corresponding to $\beta_1=86$, $\beta_2=91$, and $\beta_3=163$ hours. The resulting duration rate matrix is 

$$
\textbf{Q}=\begin{pmatrix}
-1/86 & 1/86 & 0 \\
0 & -1/91 & 1/91 \\
1/163 & 0 & -1/163
\end{pmatrix}.
$$

With a sampling interval of 20 hours equation~\ref{eqn:ref_hmm} provides the transition probability matrices. Additionally, we assume there are four possible observation types with known emission probabilities resulting in the following reference transition and emission matrices
$$
T_{ref}=\begin{pmatrix}
0.793 & 0.185 & 0.021 \\
0.011 & 0.804 & 0.185 \\
0.103 & 0.012 & 0.885
\end{pmatrix}
\quad
E_{ref}=\begin{pmatrix}
0.1 & 0.15 & 0.65 & 0.1 \\
0.05 & 0.1 & 0.5 & 0.35 \\
0.15 & 0.7 & 0.05 & 0.1
\end{pmatrix}.
$$

\begin{figure}[t!]
    \centering
    \begin{subfigure}[t]{0.45\textwidth}
        \centering
        \includegraphics[width=\linewidth]{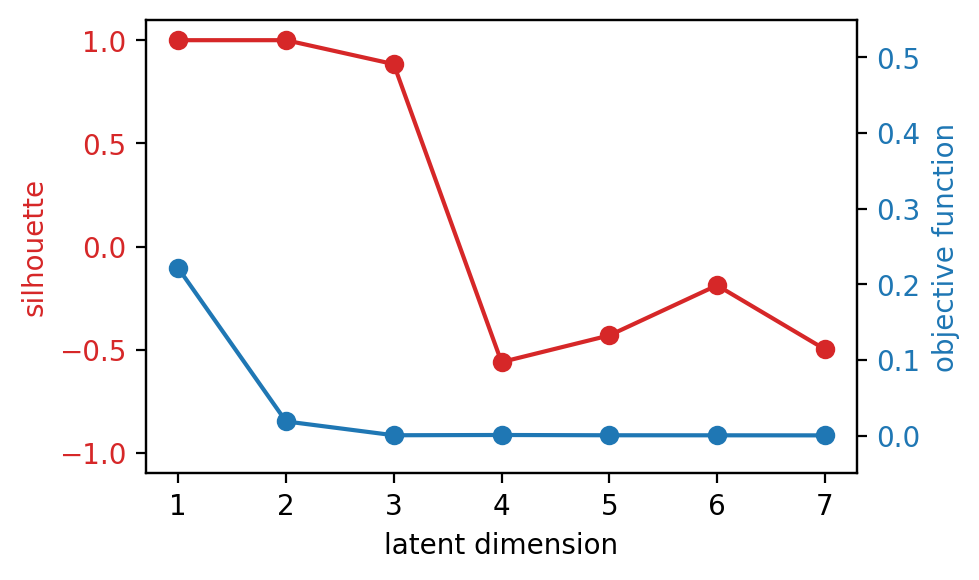}
        \caption{Silhouette scores and objective function values for each candidate latent dimension.}
        \label{fig:serial2020_k}
    \end{subfigure}%
    ~ 
    \begin{subfigure}[t]{0.45\textwidth}
        \centering
        \includegraphics[width=\linewidth]{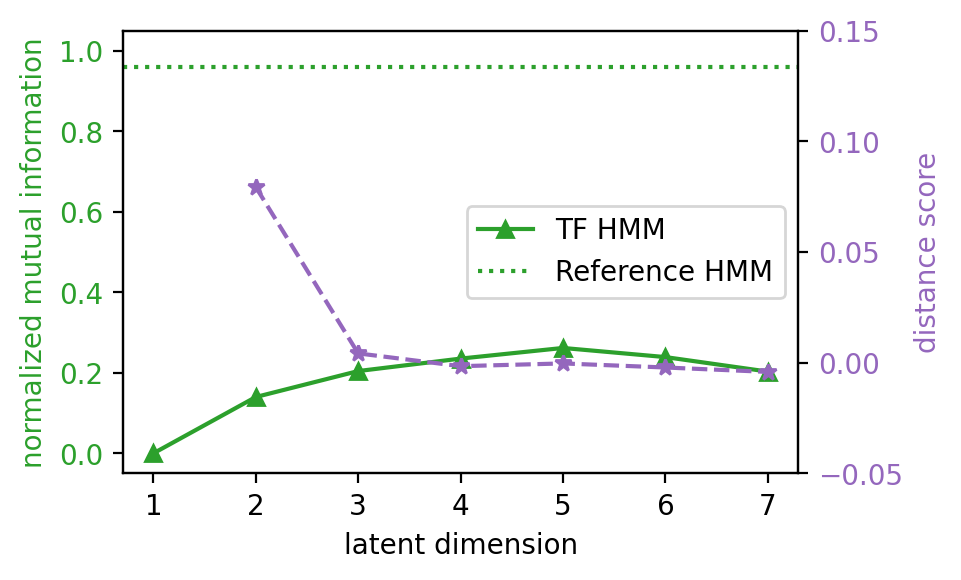}
        \caption{Normalized mutual information and distance scores for each candidate latent dimension.}
        \label{fig:serial2020_evaluation}
    \end{subfigure}
    \caption{Tensor factorization parameter selection and evaluation for simple strictly sequential process.}
    \label{fig:serial2020}
\end{figure}
Figure~\ref{fig:serial2020_k} shows the silhouette scores and objective function values for each candidate latent dimension. We see that our stability criteria clearly identifies $3$ hidden states for the HMM with a silhouette score close to 1.0 and objective value close to 0.0. The tensor factorization emission and transition matrices for $k=3$ are
$$
T_{TF}=\begin{pmatrix}
0.809 & 0.168 & 0.023 \\
0.011 & 0.787 & 0.202 \\
0.093 & 0.021 & 0.886
\end{pmatrix}
\quad
E_{TF}=\begin{pmatrix}
0.098 & 0.149 & 0.641 & 0.113 \\
0.046 & 0.095 & 0.496 & 0.363 \\
0.15 & 0.7 & 0.05 & 0.099
\end{pmatrix}.
$$

The distance score in Figure~\ref{fig:serial2020_evaluation} at each latent dimension indicates if there is an advantage of that latent dimension over the previous latent dimension. We see a large positive value at three, indicating it is advantageous over a latent dimension of two, and a value close to 0.0 at a latent dimension of four, indicating that a latent dimension of four is not significantly better than three. This aligns with the silhouette selection criteria. The normalized mutual information is relatively high at three, but peaks at a latent dimension of five. The correct selection of a latent dimension based on the silhouette and relative error demonstrates our ability to construct minimal HMMs from a theoretical process model with no parallelism.

\subsection{Synthetic Example with Process Model Activities That Do Not Lead to a Minimal HMM}

Figure~\ref{fig:concurrent_pm} depicts a process model with two activities in series followed by two activities in parallel. With the parallelism, there are 5 hidden states in the reference HMM corresponding to \{Activity 1\}, \{Activity 2\}, \{Activity 3 and 4\}, \{Activity 3\}, and \{Activity 4\}. With this process model, the reference HMM has two possible states following the state corresponding to \{Activity 3 and 4\}. Either Activity 3 will end first, so the state will change to \{Activity 4\} or vice versa. Each simulation of the process model will have different durations of each activity resulting on both paths being taken. The assigned mean durations for each activity were $\beta_1=86$, $\beta_2=91$, $\beta_3=163$, and $\beta_4=100$ hours resulting in the duration rate matrix, 

$$
\textbf{Q}=\begin{pmatrix}
-1/86 & 1/86 & 0 & 0 & 0 \\
0 & -1/91 & 1/91 & 0 & 0 \\
0 & 0 & -1/163-1/100 & 1/163 & 1/100 \\
1/163 & 0 & 0 & -1/163 & 0 \\
1/100 & 0 & 0 & 0 & -1/100
\end{pmatrix}\;.
$$

The transition probability matrix corresponding to sampling at discrete intevals of 20 hours for the repeating process model and the known emission probability matrix are 
$$
T_{ref}=\begin{pmatrix}
0.793 & 0.185 & 0.020 & 0.001 & 0.001 \\
0.001 & 0.803 & 0.168 & 0.011 & 0.017 \\
0.021 & 0.001 & 0.724 & 0.100 & 0.154 \\
0.103 & 0.011 & 0.001 & 0.885 & 0.000 \\
0.161 & 0.019 & 0.001 & 0.000 & 0.819
\end{pmatrix}
\quad
E_{ref}=\begin{pmatrix}
0.1 & 0.15 & 0.65 & 0.1 \\
0.05 & 0.1 & 0.5 & 0.35 \\
0.15 & 0.7 & 0.05 & 0.1 \\
0.15 & 0.7 & 0.05 & 0.1 \\
1 & 0 & 0 & 0
\end{pmatrix}\;.
$$

\begin{figure}[t!]
    \centering
    \begin{subfigure}[t]{0.45\textwidth}
        \centering
        \includegraphics[width=\linewidth]{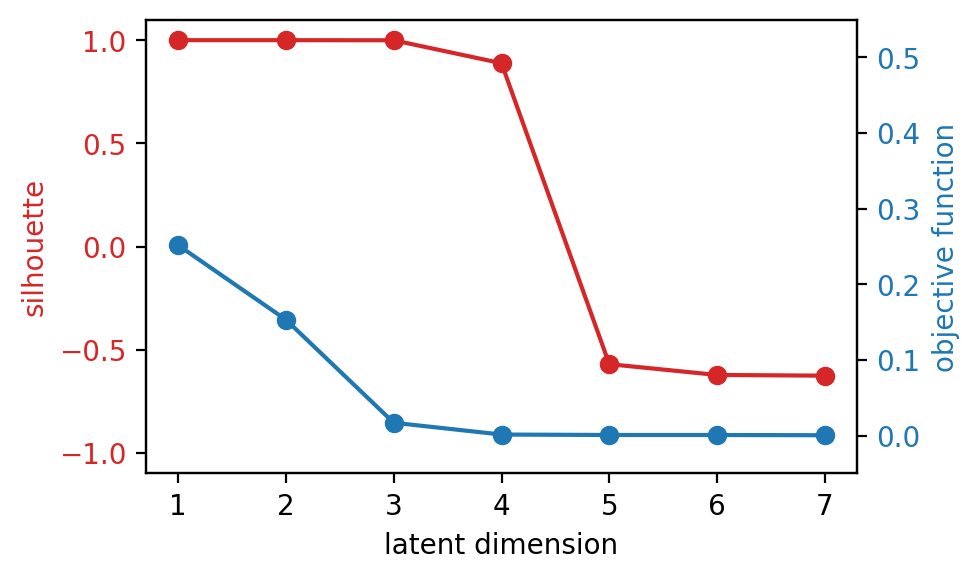}
        \caption{Silhouette scores and objective function values for each candidate latent dimension.}
        \label{fig:concurrent2020_k}
    \end{subfigure}%
    ~ 
    \begin{subfigure}[t]{0.45\textwidth}
        \centering
        \includegraphics[width=\linewidth]{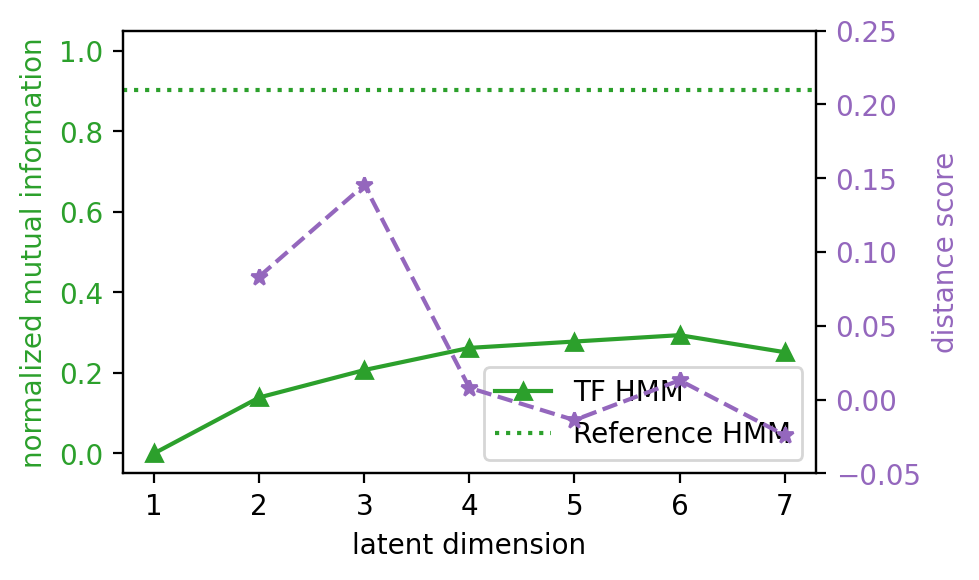}
        \caption{Normalized mutual information and distance scores for each candidate latent dimension.}
        \label{fig:concurrent2020_evaluation}
    \end{subfigure}
    \caption{Tensor factorization parameter selection and evaluation for concurrent process.}
    \label{fig:concurrent2020}
\end{figure}

Our TF selection criteria identifies four instead of five latent dimensions. This is seen in Figure~\ref{fig:concurrent2020_k} with a high silhouette score and low relative error at $k=4$. For $k=4$ the TF transition and emission matrices are

$$
T_{TF}=\begin{pmatrix}
0.755 & 0.21 & 0.033 & 0.002 \\
0.05 & 0.739 & 0.2 & 0.011 \\
0.063 & 0.022 & 0.879 & 0.036 \\
0.147 & 0.026 & 0.008 & 0.82
\end{pmatrix}
\quad
E_{TF}=\begin{pmatrix}
0.102 & 0.144 & 0.651 & 0.102 \\
0.042 & 0.093 & 0.495 & 0.37 \\
0.15 & 0.701 & 0.05 & 0.099 \\
0.994 & 0.003 & 0.001 & 0.002
\end{pmatrix}.
$$
The distance score in Figure~\ref{fig:concurrent2020_evaluation} also suggests the latent dimension is four. This is the most suitable selection, since at four the distance score is slightly positive, and at five it is negative. Interestingly the normalized mutual information continues to grow beyond four and peaks at six. By identifying a smaller HMM than the reference HMM, this validates that our selection criteria can find minimal HMMs of process models with parallelism.

In this case, the discrepancy between the reference and TF HMMs latent dimensions is due to the parallelism and the emission probabilities. From the reference emission probabilities, we immediately see that the emission probabilities between the combinations of process model activities, \{Activity 3 and 4\} and \{Activity 3\} are identical. This equivalence induces a degeneracy in the reference HMM allowing for an equivalent HMM to be constructed with one fewer hidden states. This is identified automatically in our coupled nonnegative tensor factorization with model selection to get a minimal HMM with $k=4$. The is seen in the minimal HMMs third state approximating the emissions of both the reference HMMs states three and four. While this example has an easily identifiable explanation for the reference HMM not being minimal, with real data this almost never the case.

\subsection{Real Life Example: HMM of a Dutch Bank Loan Application Process Model}
To demonstrate our techniques on real data, we looked at a set of event logs for a loan approval process for a bank in the Netherlands (for the bank data see \cite{moreira2018process} and references therein). The data set contains event logs for 13,087 loan applications. Each event log contains the hidden process model activities (reviewing application at each ti, calling customer for information, etc.). Additionally, each log contains IDs and timestamps indicating when each bank employee is actively accessing the application. We sample the the sequence of the 48 unique employee IDs accessing each application as our our observable resource. The final loan outcome for each log is also included (e.g., acceptance, rejection). 

The true loan application process is very complicated, since there are different paths a loan might take through the process (e.g. it might get accepted or rejected). Real procedures do not necessarily adhere to an acyclic process as it is possible to return to an activity previously completed. Also, there are long time gaps (e.g. breaks, evenings, waiting for customer response, etc.) between activities and within activities, and there are activities that show up in some runs of the process but not others (e.g. sometimes an application is investigated for fraud, sometimes it is not). In order to simplify the modeling, we consider a subset of the loan application data. First, we only consider loans that made it to final approval (there are 2,446 out of 13,087 such applications).

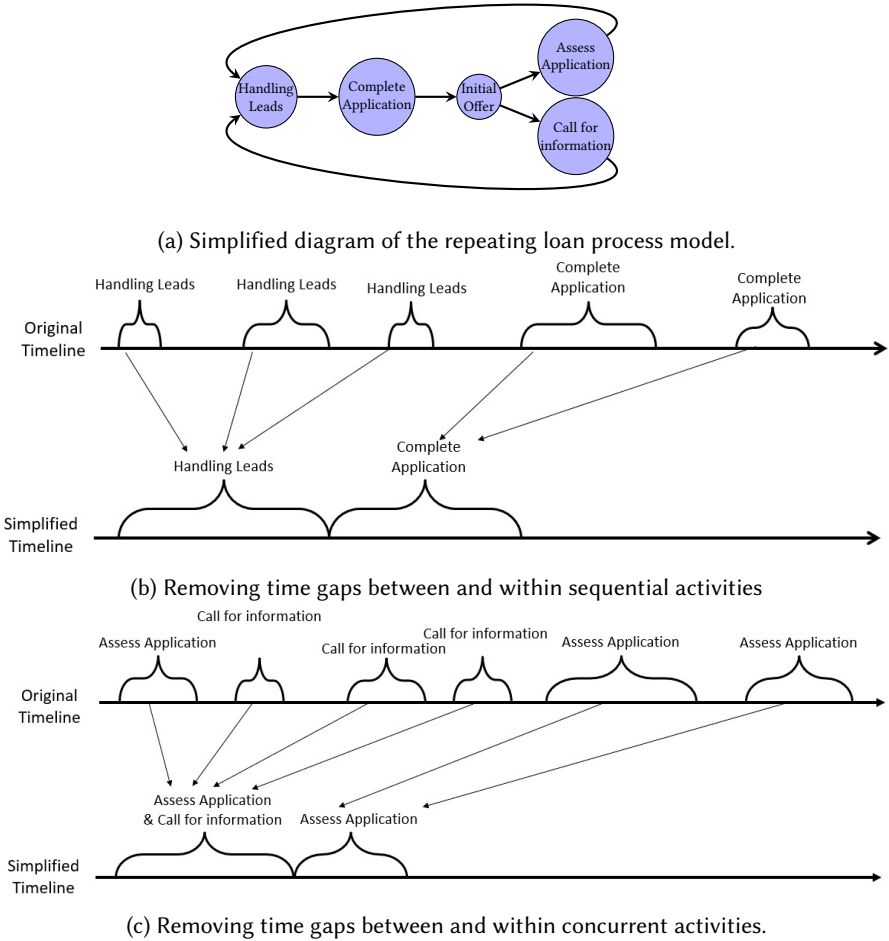
\begin{figure}
    \centering
    \begin{subfigure}[t]{\textwidth}
        \centering
    \begin{tikzpicture}[node distance=.0cm and 1cm, scale=0.55, every node/.append style={transform shape}]
    
    \clip (-1,-2.75) rectangle (9.75,2.5);
    
    \node (1) [bluecirc] {Handling\\ Leads};
    \node (2) [bluecirc, right=of 1] {Complete\\ Application};
    \node (3) [bluecirc, right=of 2] {Initial\\ Offer};
    \node (4) [bluecirc, above right=-.1cm and 1.3cm of 3] {Assess\\ Application};
    \node (5) [bluecirc, below right=-.1cm and 1.3cm of 3] {Call for\\ information};
    \draw [arrow] (1) -- (2);
    \draw [arrow] (2) -- (3);
    \draw [arrow] (3) -- (4);
    \draw [arrow] (3) -- (5);
    \draw [arrow] (4) to[out=35, in=145, looseness=.75] (1);
    \draw [arrow] (5) to[out=-35, in=-145, looseness=.75] (1);
    \end{tikzpicture}
    \caption{Simplified diagram of the repeating loan process model.}
    \label{LoanProcess}
    \end{subfigure}\\
    \begin{subfigure}[t]{\textwidth}
        \centering
        \includegraphics[width=0.85\textwidth]{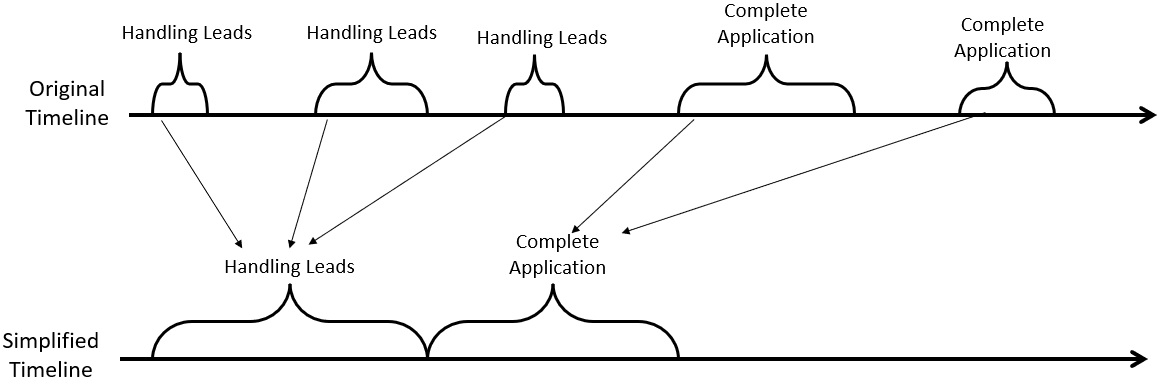}
        \caption{Removing time gaps between and within sequential activities}
        \label{TimeGaps1}
    \end{subfigure}\\
    \begin{subfigure}[t]{\textwidth}
        \centering
        \includegraphics[width=0.85\textwidth]{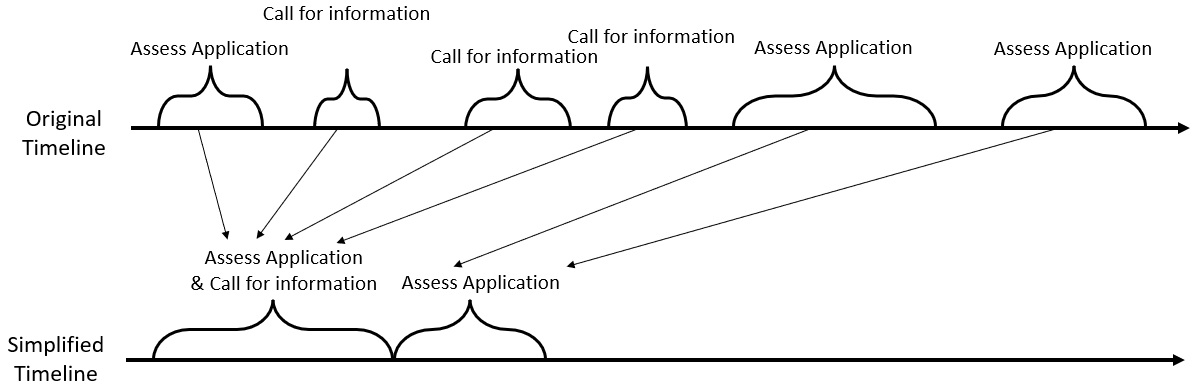}
        \caption{Removing time gaps between and within concurrent activities.}
        \label{TimeGaps2}
    \end{subfigure}
    \caption{Diagrams showing the simplified load process and the necessary steps to clean the data.}
\end{figure}

 
Of the approved applications, we only consider applications that adhere to the process model diagram in Figure~\ref{LoanProcess}. The first activity, Handling Leads, refers to bank agents determining whether the applicant is someone the bank already had a lead on. In the second activity, Complete Application, the bank employees ensure the completeness of the loan application, seek out additional information from the customer, and complete any parts of the application required on behalf of the bank. The Initial Offer, is the bank making an initial offer to patrons. After an initial offer two activities can happen simultaneously. If the customer is not satisfied with the initial offer, they can ask the employees to reassess the application. While this reassessment is happening, employees may call the customer for additional information. Of the 2,446 approved loans, 347 cases match this ordering of activities. 

We further simplify the process by deleting the time gaps between and within activities. These gaps pertain to when workers were busy with other tasks, nights, weekends, holidays, and other instances when an application was not actively being handled. By removing these gaps, the resulting timelines indicate the working-hours necessary for an application to be processed. In the true process, there are long gaps between when an activity is finished and when the next one is started, we simplify this and assume that the activities happen one right after the other. In the true process, part of an activity is completed by one employee, then there is usually a time gap and then another part of the activity is completed by another employee. We eliminate this time gap between different parts of the same activity. The total duration spent in each activity is then the sum of the times between the start and end of each part of each activity. These simplifications are illustrated for sequential activities in Fig. \ref{TimeGaps1}. 

For activities running in parallel, like the last two activities of this process, the simplification is a little more complex. We again take the total duration of each of the activities running in parallel to be the sum of the gaps between each start and end time, but we also determine when these activities would start relative to one another. In the true process there is a gap between the start of one activity and the start of the other activity that runs parallel with it. We simplify the model by assuming the activities start at the same time. The activity that finishes first is the one with shorter duration as determined previously. The simplification for parallel activities is depicted in Fig. \ref{TimeGaps2}.

The scarce real data of only 347 data logs, necessitates the generation of additional data using Simian. From the 347 simplified timelines with gaps removed, we compute the expected duration of each activity from the data. Additionally, we use the probability that each of the 48 employeed accessef the application during each process activity as our probability that each employee will be observed. These parameters are used with Simian to simulate runs and sample the observed employee access at 15 minute intervals. The reference transition and emission matrices are also generated from this data. Using the abundant simulated data, we empirically constructed the joint probability tensor and joint probability matrices and apply our technique to construct a minimal HMM.

\begin{figure}[t!]
    \centering
    \begin{subfigure}[t]{0.45\textwidth}
        \centering
        \includegraphics[width=\linewidth]{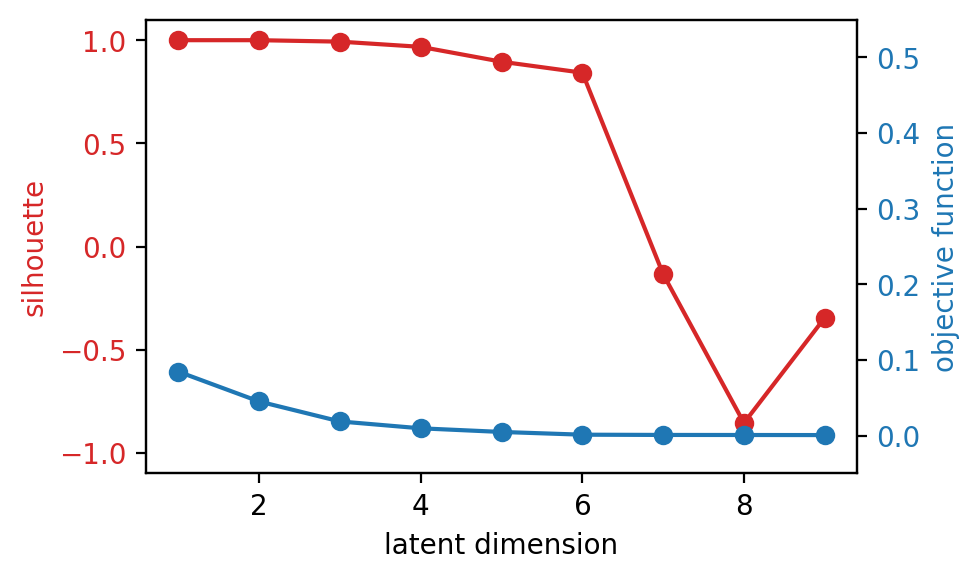}
        \caption{Silhouette scores and objective function values for each candidate latent dimension.}
        \label{fig:banking_k}
    \end{subfigure}%
    ~ 
    \begin{subfigure}[t]{0.45\textwidth}
        \centering
        \includegraphics[width=\linewidth]{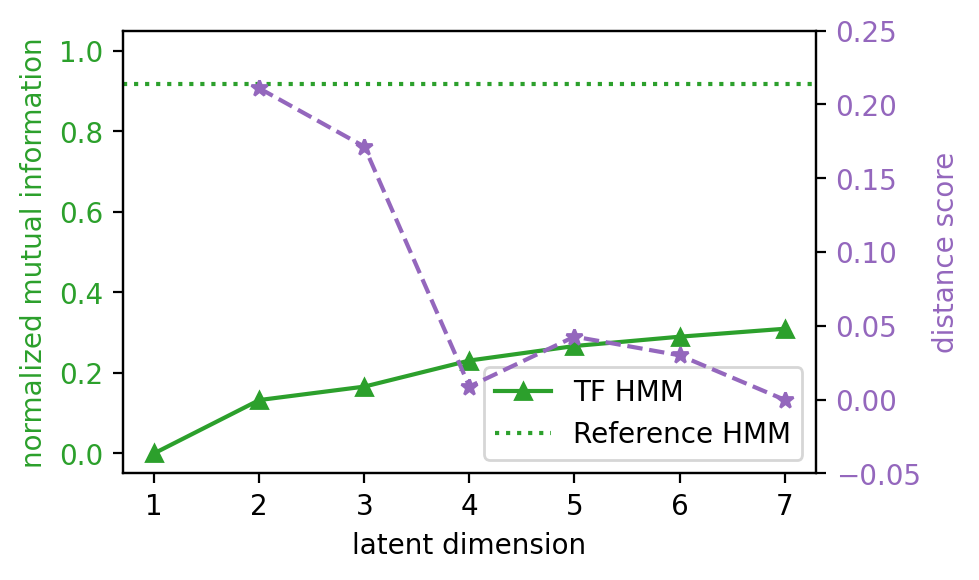}
        \caption{Normalized mutual information and distance scores for each candidate latent dimension.}
        \label{fig:banking_evaluation}
    \end{subfigure}
    \caption{Tensor factorization parameter selection and evaluation for banking process model.}
    \label{fig:banking}
\end{figure}

The silhouette scores and relative errors in Figure~\ref{fig:banking_k} clearly indicate six identifiable features, the same number of latent features as of the reference HMM. The reference and tensor factorization transition matrices are,
$$
T_{ref}=\begin{pmatrix}
0.578 & 0.323 & 0.084 & 0.013 & 0.001 & 0.001 \\
0.001 & 0.6 & 0.314 & 0.075 & 0.004 & 0.007 \\
0.004 & 0.001 & 0.629 & 0.301 & 0.024 & 0.041 \\
0.029 & 0.005 & 0.001 & 0.67 & 0.111 & 0.184 \\
0.103 & 0.027 & 0.004 & 0.001 & 0.865 & 0.0 \\
0.172 & 0.045 & 0.008 & 0.001 & 0.0 & 0.774
\end{pmatrix}
$$

$$
T_{TF}=\begin{pmatrix}
0.596 & 0.307 & 0.089 & 0.004 & 0.002 & 0.004 \\
0.002 & 0.597 & 0.31 & 0.005 & 0.031 & 0.054 \\
0.018 & 0.002 & 0.64 & 0.086 & 0.107 & 0.147 \\
0.139 & 0.04 & 0.03 & 0.285 & 0.377 & 0.129 \\
0.058 & 0.036 & 0.001 & 0.083 & 0.814 & 0.009 \\
0.071 & 0.06 & 0.003 & 0.055 & 0.074 & 0.737 
\end{pmatrix}\;.
$$

The tensor and matrix factorization errors for $k=6$ are
\begin{equation}
    \begin{aligned}
    \frac{1}{2}\Vert \mathcal{T} - \Brackets{A,B,C}\Vert_F^2 & \approx 0.0013,\\
    \frac{\rho}{2} \Vert M - B D B^\top \Vert_F^2 & \approx 0.0034.
    \end{aligned}
\end{equation}
The emission matrices for this example are not shown due to their significant number of observed states.
The selection of $k=6$ agrees with the distance score analysis in Figure~\ref{fig:banking_evaluation} with a positive value at six, and approximately zero at seven. The normalized mutual information curve reported in Figure~\ref{fig:banking_evaluation} was computed using the real event logs of observed employee access and process activities. While normalized mutual information is higher at seven, at $k=6$ the normalized mutual information of approximately 0.3 shows a significant correlation between the hidden states predicted by the Viterbi algorithm, and the activities underway in the real data at each 15 minute interval. This exhibits that augmentation of scarce real data with SME informed process models can provide a way to relate the real data to a constructed minimal HMM. This relation is useful, as in cases where the reference HMM has many more states than a minimal HMM, it suggests that closer scrutiny of the process model is necessary to potentially eliminate unnecessary activities. Additionally, the generation of minimal HMMs for competing process model allows for evaluation of which process model fits the scarce real data more accurately though the application of the forward algorithm.

\section{Conclusion}
Here we introduce a new unsupervised machine learning method based on nonnegative tensor factorization for building  minimal HMMs based on a theoretical process model constructed from domain expert estimations. Our new method is based on our technique for model selection in NMF and NTF, and gives us the capability to identify the minimal number of states needed in an HMM resulting from a process model. In the future, this work could furnish the ability to inform SMEs of unnecessary complexity in their process models, and to compare competing process models. In this work we demonstated our ability to determine the unknown number of the hidden states in HMM as well to work with theoretical process models with sequential and parallel activities. 
The minimal HMMs built in this way allows us to answer questions of interest using domain expertise and scarce observational data.

\begin{acks}
This research was funded by DOE National Nuclear Security Administration  (NNSA)  -  Office  of  Defense  Nuclear  Non-proliferation  R\&D and by U.S. Department of Energy National Nuclear Security Administration under Contract No. DE-AC52-06NA25396 through LANL laboratory support.
\end{acks}

\section*{Conflict of interest}
The authors declare that they have no conflict of interest.

\bibliographystyle{ACM-Reference-Format}
\bibliography{references}

\end{document}